\documentclass[letterpaper]{article} 
\usepackage{aaai25}
\usepackage{multirow}
\usepackage{array}
\usepackage{amsmath}
\usepackage{amssymb}
\usepackage{bm}
\usepackage{subfig}
\usepackage{booktabs} 
\usepackage{times}  
\usepackage{helvet}  
\usepackage{courier}  
\usepackage[hyphens]{url}  
\usepackage{graphicx} 
\usepackage{natbib}  
\usepackage{caption} 
\usepackage{algorithm}
\usepackage{algorithmic}
\usepackage{newfloat}
\usepackage{listings}

\DeclareCaptionStyle{ruled}{labelfont=normalfont,labelsep=colon,strut=off} 
\floatstyle{ruled}
\newfloat{listing}{tb}{lst}{}
\floatname{listing}{Listing}

\title{Hiformer: Hybrid Frequency Feature Enhancement Inverted Transformer for Long-Term Wind Power Prediction}
\author{
    Chongyang Wan,\textsuperscript{\rm 1}
    Shunbo Lei,\textsuperscript{\rm 1}
    Yuan Luo\textsuperscript{\rm 1}\\
}
\affiliations{
    \textsuperscript{\rm 1}School of Science and Engineering, The Chinese University of Hong Kong, Shenzhen, 
Guangdong, 518172, P.R. China \\
    chongyangwan@link.cuhk.edu.cn, leishunbo@cuhk.edu.cn, luoyuan@cuhk.edu.cn

}

\usepackage{bibentry}

\begin{document}

\maketitle

\begin{abstract}
The increasing severity of climate change necessitates an urgent transition to renewable energy sources, making the large-scale adoption of wind energy crucial for mitigating environmental impact. However, the inherent uncertainty of wind power poses challenges for grid stability, underscoring the need for accurate wind energy prediction models to enable effective power system planning and operation. While many existing studies on wind power prediction focus on short-term forecasting, they often overlook the importance of long-term predictions. Long-term wind power forecasting is essential for effective power grid dispatch and market transactions, as it requires careful consideration of weather features such as wind speed and direction, which directly influence power output. Consequently, methods designed for short-term predictions may lead to inaccurate results and high computational costs in long-term settings. To adress these limitations, we propose a novel approach called Hybrid Frequency Feature Enhancement Inverted Transformer (Hiformer). Hiformer introduces a unique structure that integrates signal decomposition technology with weather feature extraction technique to enhance the modeling of correlations between meteorological conditions and wind power generation. Additionally, Hiformer employs an encoder-only architecture, which reduces the computational complexity associated with long-term wind power forecasting. Compared to the state-of-the-art methods, Hiformer: (i) can improve the prediction accuracy by up to 52.5\%; and (ii) can reduce computational time by up to 68.5\%.
\end{abstract}

\section{Introduction}
Over recent decades, the rapid consumption of fossil fuels has exacerbated climate warming, making the transition to renewable energy increasingly urgent. As a clean, safe and renewable energy source, wind power serves as an effective alternative to fossil fuels, significantly reducing carbon emissions \citep{zhao2024spatial}. Hence, we have witnessed a large-scale wind power grid connections in reality \citep{TAWN2022111758}. However, unlike traditional power station, the output of wind power plants is highly dependent on weather parameters such as wind speed, wind direction and temperature. This dependence introduces significant volatility and intermittency in wind power production, making grid systems highly fragile and challenging to manage \citep{wang2021review,liu2024wind}. 

To reduce potential supply and load imbalances in the power grid caused by the uncertainty of wind power, accurate wind power prediction is crucial. This enables grid operators to flexibly adjust the output of each generator and energy storage unit in advance \citep{10497893}. Existing studies primarily focus on designing deep learning models to explore the spatial-temporal correlations of wind power data (see Related Work Section for details). However, these methods often fail to produce accurate long-term predictions (i.e., one day in advance) due to the highly non-stationary nature of wind power. As shown in Figure 1, unlike traffic or solar data, which exhibit more predictable patterns, wind power patterns change unpredictably over time. Factors such as seasonal shifts, climate variations and other meteorological influences can cause significant changes in wind patterns, making it difficult for models that only consider spatial-temporal data to accurately capture and predict these long-term variations \citep{wang2023weagan}. 

The issues of incorporating such weather impact into wind power prediction have recently been studied in \citep{he2022combined, liu2023itransformer}. However, these work neglect the computational complexity problem. As a result, their methods require substantial computational time, rendering them impractical for real-world wind power systems \citep{du2020hybrid,wang2021review}.

\begin{figure}[t]
\centering
\includegraphics[width=0.47\textwidth]{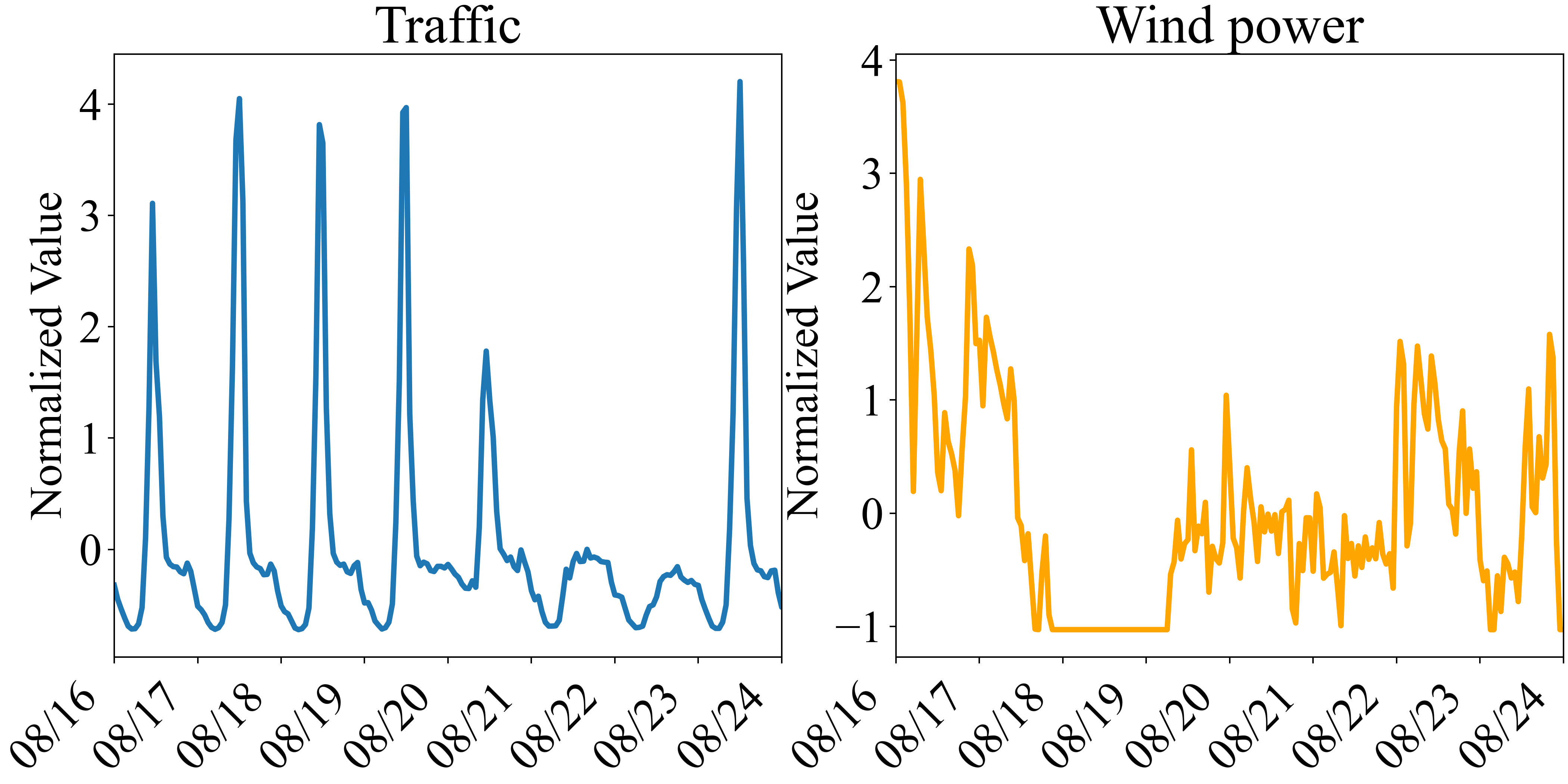} 
\caption{Traffic and wind power data examples from Aug-16 to Aug-24.}
\label{Figure1}
\end{figure}

Designing a model for long-term wind power prediction with low computational complexity is non-trivial and has not been done before. The complexity arises from the non-stationary nature of wind power, which requires additional techniques to handle the unpredictable wind patterns influenced by varying weather conditions. Additionally, widely used techniques, such as transformer-based architectures, which are effective at capturing feature correlations in data, often suffer from performance degradation and computational inefficiency when applied to explore the spatial-temporal dependencies between wind power generation and weather features. Hence, we cannot simply combine existing methods to achieve accurate predictions within a reasonable computational time.

Against this background, we propose a Hybrid Frequency Feature Enhancement Inverted Transformer (Hiformer) for long-term wind power prediction. By carefully incorporating the Variational Mode Decomposition (VMD) technique within an encoder-only architecture, Hiformer effectively addresses the non-stationary nature of wind power caused by varying weather conditions. Furthermore, Hiformer adopts inverted multi-attention mechanisms \citep{liu2023itransformer} to prevent computational explosion during the prediction process.

This paper advances the state-of-the-art in the following ways. First, Hiformer is the first wind power prediction model to integrate VMD with weather feature extraction methods. This integration enhances Hiformer’s ability to predict wind power generation that are highly influenced by unpredictable weather conditions. Second, by carefully designing an encoder-only structure with an inverted attention mechanism, Hiformer achieves computational efficiency, making it feasible for practical wind power systems. Finally, our empirical studies show that by considering the non-stationary nature of wind power and applying the VMD technique together with weather feature extraction technique to mitigate the impact of weather conditions, Hiformer improves prediction accuracy by up to 52.5\%. Additionally, compared to state-of-the-art models, Hiformer’s encoder-only structure with an inverted attention mechanism reduces computational time by up to 68.5\%. 

\section{Related Work}
To capture the non-linear and fluctuating temporal correlations in wind power data, researchers have adopted deep learning techniques \citep{giebel2017wind}. For example, \citet{kisvari2021wind} employed Recurrent Neural Networks (RNNs) to model complex temporal relationships in wind power data by maintaining a memory of past states. \citet{liu2024wind} developed a Long Short-Term Memory (LSTM) model that selectively retains and forgets information over long sequences, thereby enabling more accurate wind power predictions. Given the highly variable and unpredictable nature of wind power data, researchers have also applied signal decomposition techniques in deep learning models to further enhance prediction accuracy \citep{jiang2024applicability}. For instance, \citet{abedinia2020improved} applied Empirical Mode Decomposition (EMD) method to decompose wind power data, followed by the use of a back propagation neural network for prediction. \citet{sun2019short} employed the Variational Mode Decomposition (VMD) method to achieve short-term wind power prediction. However, these methods often overlook the influence of spatial information on prediction accuracy. Note that wind turbines possess spatial characteristics within a wind farm, which can significantly impact prediction accuracy \citep{zhao2024spatial}. For example, wind turbines located downwind in a wind farm are significantly affected by the wake of the adjacent upwind wind turbines, resulting in reduced power generation \citep{mittelmeier2017monitoring,wang2021review}. 

To incorporate turbine spatial information in wind power prediction, a commonly used method is integrating graph structures of wind turbines into the prediction model \citep{qiu2024novel}. For example, \citet{wang2022dynamic} proposed an ultra-short-term wind farm cluster power forecasting method that effectively improves prediction accuracy. \citet{yu2020superposition} developed a superposition graph neural network for extracting spatial features. However, a common limitation of these methods is that they only consider the correlations between wind turbines, neglecting the impact of weather conditions on wind power generation. Given that wind power generation is highly dependent on weather features such as wind speed and direction, these methods fall short in providing accurate long-term predictions.

To address this, many studies incorporate weather features into spatial-temporal prediction models using attention mechanisms to capture multivariate relationships \citep{zheng2020gman}. For example, \citet{he2022combined} achieved long-term wind power prediction under varying weather conditions. \citet{liu2023itransformer} modified the attention module in the traditional transformer model \citep{vaswani2017attention} to focus on the correlations between weather features and wind power, enabling long-term predictions across multiple turbines. Although these studies have made some progress, they require substantial computing time to calculate the complex relationships between weather features and wind power, limiting their practicality for real-world applications.

\section{Hybrid Frequency Feature Enhancement Inverted Transformer}
In this section, we first present the mathematical definition of the wind power prediction problem. Following this, we offer a detailed description of our proposed Hiformer.

\subsection{Problem Definition}
Our goal is to predict the wind power generation $\bm{\hat{X}} =(\hat{X}_{t_{P+1}},\hat{X}_{t_{P+2}},...,\hat{X}_{t_{P+Q}})\in \mathbb{R} ^{Q\times N}$ for next $Q$ time steps, given $N$ turbines' wind power generation $\bm{X} =(X_{t_{1}},X_{t_{2}},...,X_{t_{P}})\in \mathbb{R} ^{P\times N}$ and weather data $\bm{W} =(W_{t_{1}},W_{t_{2}},...,W_{t_{P}})\in \mathbb{R} ^{P\times N\times C}$ for $P$ historical time steps. Here, $C$ is the number of weather features (e.g., wind speed, wind direction and ambient temperature).

In order to characterize the spatial correlations among power outputs of $N$ wind turbines, we represent a weighted undirected graph $G=(V,E,A)$, where $V$ is the set of locations of $N$ wind turbines, $E$ is the set of edges between turbines and $A\in \mathbb{R} ^{N\times N}$ is a weighted adjacency matrix with $A_{v_{i},v_{j}}\in [0,1], v_{i},v_{j}\in V$ denoting the proximity between vertices $v_{i}$ and $v_{j}$.
\begin{figure}[t]
\begin{minipage}[b]{.48\linewidth}
    \centering
    \subfloat[][Overall structure of Hiformer]{\label{Figure2_1}\includegraphics[width=1.0\linewidth]{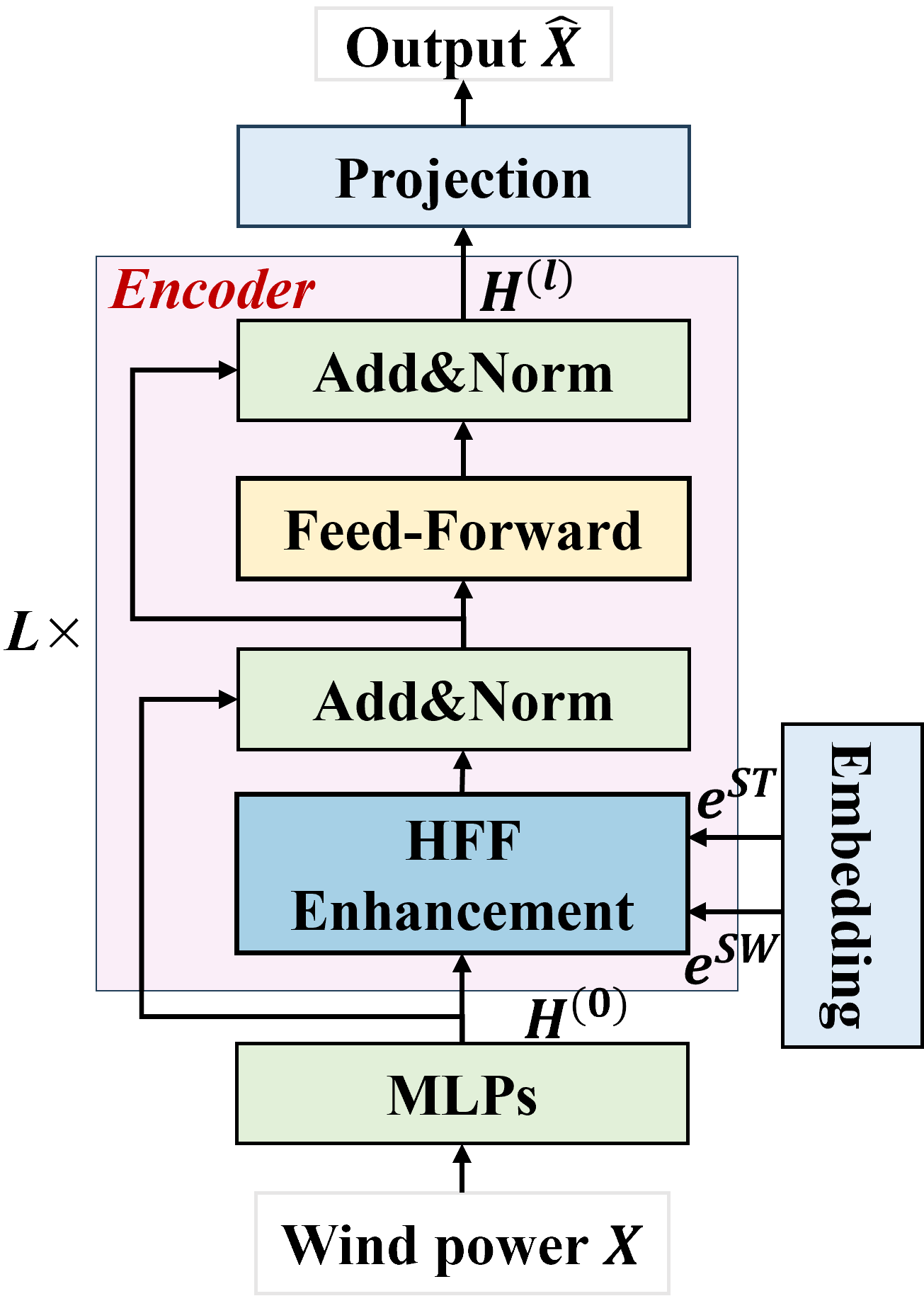}}
\end{minipage} 
\hspace{0.3cm}
\begin{minipage}[b]{.46\linewidth}
    \centering
    \subfloat[][Embedding Blocks]{\label{Figure2_2}\includegraphics[width=1.0\linewidth]{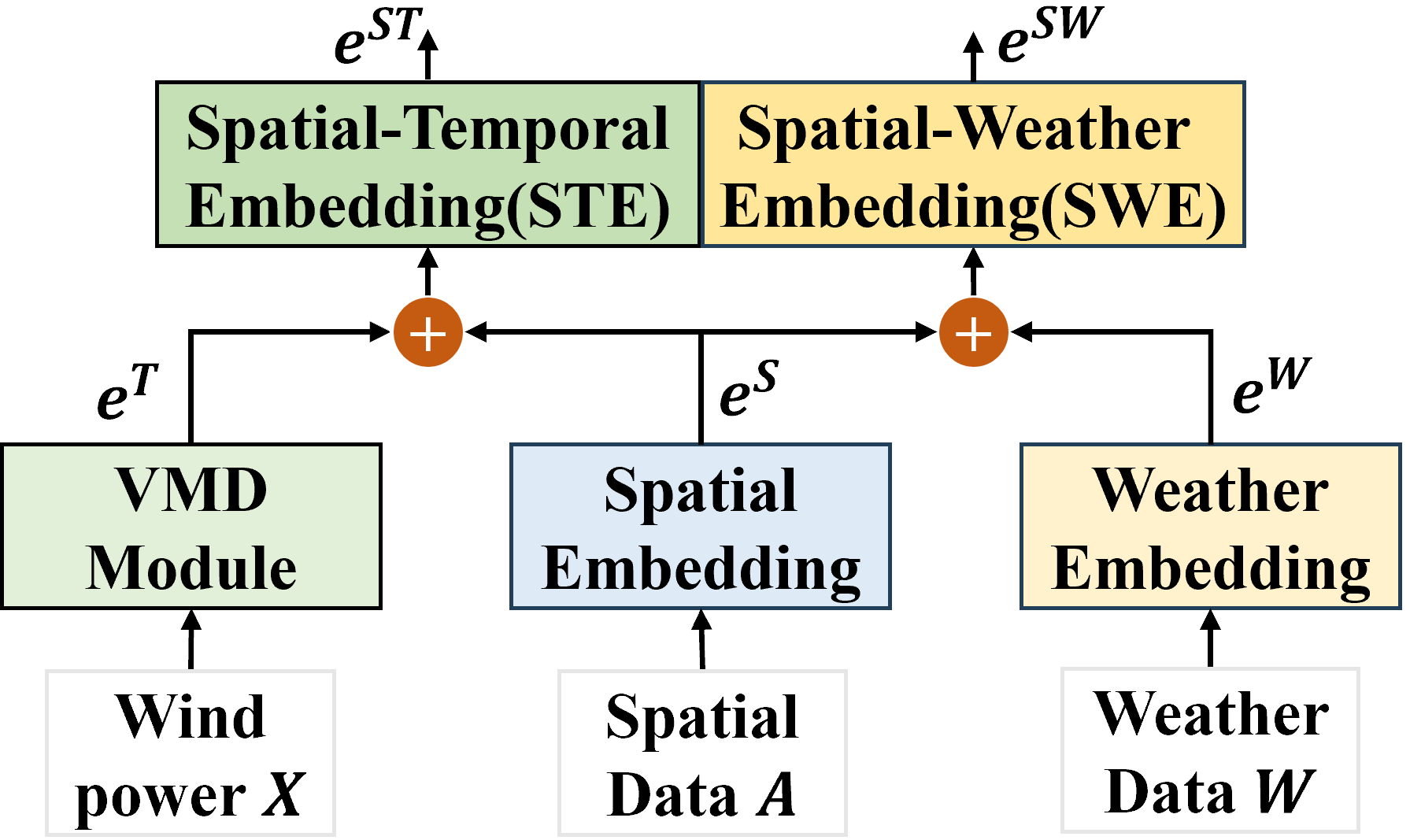}}

    \subfloat[][HFF Enhancement Block]{\label{Figure2_3}\includegraphics[width=1.0\linewidth]{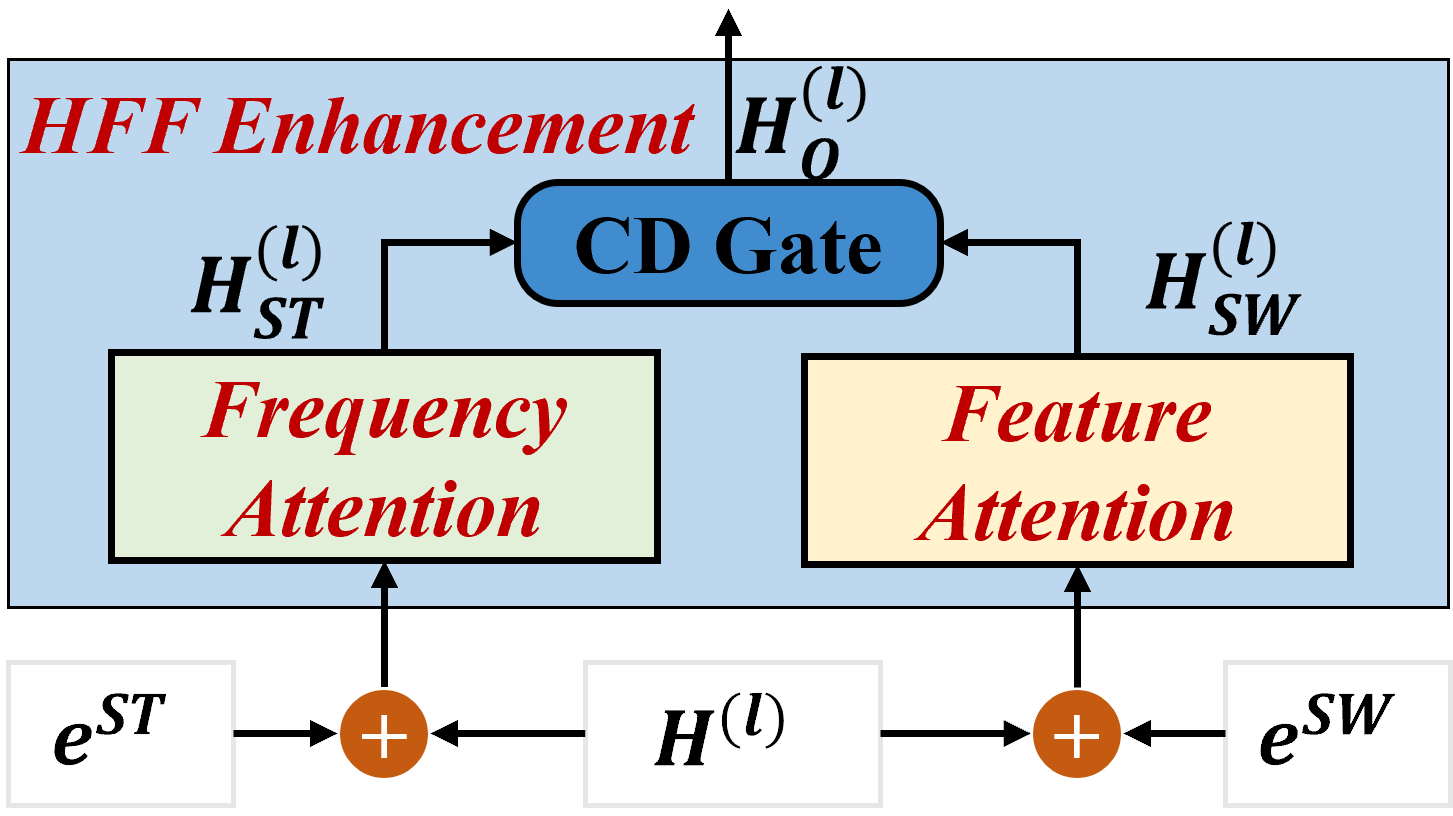}}
\end{minipage}
\caption{The framework of Hiformer.}
\label{Figure2}
\end{figure}
\subsection{Model Overview}
As shown in Figure 2(a), Hiformer employs an encoder-only structure \citep{vaswani2017attention}. In more details, the model begins with an Embedding block to extract temporal, spatial, and weather information from the raw data. This is followed by the Hybrid Frequency Feature Enhancement (HFF) block, which enhances Hiformer's ability to capture the spatial-temporal correlations between wind power generation and weather features. To prevent overfitting, we employ an Add\&Norm layer, and a Feed-Forward network is used to strengthen the temporal representations of wind power. Finally, a Projection block decodes the encoder's output, producing the final prediction results. Detailed explanations of these components are provided below.

\subsection{Embedding}
As shown in Figure 2(b), we use three embedding blocks to process wind power data $X$, spatial data $A$ and weather features data $W$. We denote the output of the VMD Module, Spatial Embedding and Weather Embedding as $e^{T}$, $e^{S}$ and $e^{W}$, respectively.
We also denote the output of Spatial-Temporal Embedding and Spatial-Weather Embedding as $e^{ST}$ and $e^{SW}$, respectively. The followings are the detailed explanations.
\begin{figure}[t]
\centering
\includegraphics[width=0.4\textwidth]{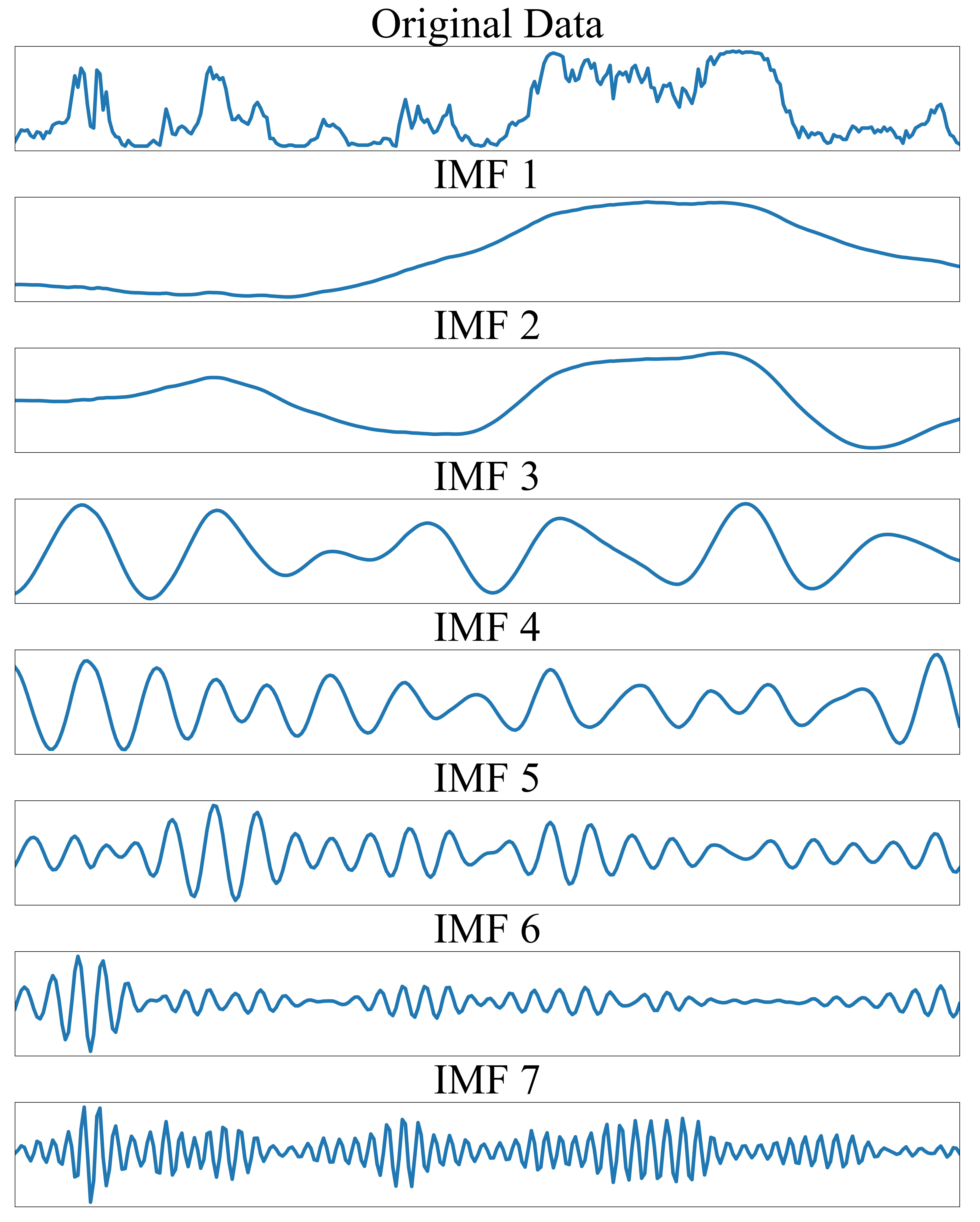} 
\caption{Original wind power data and 7 IMFs after Variational Mode Decomposition.}
\label{Figure3}
\end{figure}
\paragraph{Variational Mode Decomposition Module:}
We first employ Variational Mode Decomposition (VMD) Module to process the wind power data. VMD is a signal decomposition technique that decomposes highly volatile wind power data into $M$ amplitude-modulated-frequency-modulated (AM-FM) signals, known as the Intrinsic Mode Function (IMF) \citep{dragomiretskiy2013variational}. As shown in Figure 3, the IMF exhibits stronger periodicity compared to the original wind power data, enabling us to better capture their temporal dependencies. We denote each IMF as $u_{i}\in\left \{ u_{1},u_{2},\dots ,u_{M} \right \}$ and define it as:
\begin{equation}
    u_{i}(t)=A_{i}(t)\cos( \phi_{i} (t)),
\end{equation}
where $A_{i}(t)$ is the non-negative amplitude, $\phi_{i} (t)$ is the phase of the signal. Note that IMF $u_{i}(t)$ can be approximated as a pure harmonic signal with amplitude $A_{i}(t)$ and central frequency $\omega _{i}(t)={\phi}' _{i}(t) $.

To obtain $M$ IMFs in each turbine, we solve the following constrained variational problem:
\begin{equation}
\underset{u_{v,i},\omega _{v,i}}{\min}\left\{\sum_{i=1}^{M}\left\|\partial_{t}\left[\left(\delta\left(t\right) + \frac{j}{\pi t} \right) \ast u_{v,i} \left( t \right) \right] e^{-j\omega_{v,i}t} \right\|_{2}^{2} \right \},\nonumber
\end{equation}
\begin{equation}
\mathrm{s.t.} \sum_{i=1}^{M}u_{v,i}=f_{v}.
\end{equation}
where $u_{v,i} $ is the $i$-th IMF given the $v$-th turbine's wind power data, $\omega _{v.i}$ is $i$-th IMF's central frequency at $v$-th turbine, $\left \|  \bullet \right \| _{2}^{2} $ is the squared euclidean norm, $\partial_{t}$ is the partial derivative of the function for time $t\in \left \{ 1,2,\dots ,P \right \} $, $\delta\left(t\right)$ is the unit pulse function, $j$ is the imaginary unit, $\ast $ indicates the convolution operation and $f_{v}$ is the original wind power data at vertex $v$. 

As is common in the literature \citep{dragomiretskiy2013variational,wang2019deep}, we apply Hilbert transform to solve this problem, and obtain the $M$ IMFs vector $\left \{ u_{v,i} \right \}=\left \{ u_{v,1},u_{v,2},\dots ,u_{v,M} \right \} $ of vertex $v$. Then we combine all the IMFs together to obtain the matrix $e^{IMF}=\left \{ \left \{ u_{1,i} \right \},\left \{ u_{2,i} \right \},\dots \left \{ u_{N,i} \right \}  \right \} \in \mathbb{R} ^{P\times N \times M}$ for $N$ turbines in $P$ times steps. Similarly to \citet{liu2023itransformer}, we apply multi-layer perceptrons (MLPs) to transform IMFs vector $e^{IMF}$ into $e^{T}\in \mathbb{R} ^{D\times N\times M}$, where $D$ represents the number of correlation hidden states for $P$ time steps. This transformation can enhance the model’s ability to capture the temporal relationships between wind power and IMFs (as we will show in the Experiments Section)
\paragraph{Spatial Embedding:}
We employ node2vec \citep{grover2016node2vec}, which is commonly used in the literature \citep{hou2021short,wang2023weagan}, to capture spatial features among wind turbines in graph $G$. The input of node2vec is the weighted adjacency matrix $A\in \mathbb{R} ^{N\times N}$, and the output is $e^{node}\in \mathbb{R} ^{P\times N}$. We then apply MLPs to process $e^{node}$, ensuring that its matrix dimensions are consistent with the VMD module's output, and obtain the Spatial Embedding output $e^{S}\in  \mathbb{R} ^{D\times N}$.
\paragraph{Weather Embedding:} 
We apply MLPs to process weather feature data $\bm{W}$, enabling us to characterize the correlations between weather features and wind power generation, and obtain the transformed output $e^{W}\in \mathbb{R} ^{D\times N \times C}$, where $D$ is the correlation hidden state, $N$ is the number of turbines and $C$ is the number of weather features.
\paragraph{Spatial-Temporal Embedding:}
As shown in Figure 2(b), we design Spatial-Temporal Embedding (STE) to capture the temporal relationship between IMFs and wind power generation within a spatial context. We superimpose the spatial embedding and VMD module outputs to obtain the STE output $e^{ST}=(e^{S}+e^{T})\in \mathbb{R} ^{D\times N\times M}$, which represents the spatial-temporal correlations of $M$ IMFs across $N$ wind turbines in $D$ dimensions.
\paragraph{Spatial-Weather Embedding:}
As shown in Figure 2(b), we design Spatial-Weather Embedding (SWE) to capture the impact of weather features on wind power generation within a spatial context. We superimpose the spatial embedding and weather embedding outputs to obtain the SWE output $e^{SW}=(e^{S}+e^{W})\in\mathbb{R} ^{D\times N\times C}$, which represents the spatial-temporal correlations of $C$ weather features across $N$ turbines in $D$ dimensions.
\subsection{Hybrid Frequency Feature Enhancement Block}
As shown in Figure 2(c), HFF Enhancement block has two attention modules, where the Frequency Attention module captures the correlations between IMFs and wind power generation within a spatial context, and Feature Attention module captures the correlations between weather features and wind power generation in a spatial context. Additionally, we design a Correlation Determining Gate (CD Gate) to adaptively integrate the correlations from these attention mechanisms. We provide the details of these modules below.

As shown in Figure 2(a), we define the input of encoder as $H^{(0)}$, the number of encoder layers as $L$ and the input of $l$-th HFF Enhancement Block as $H^{(l)}$. The hidden state for vertex $v$ at correlation dimension $d$ is represented as $h_{v,d}^{(l)}$ and $H^{(l)}=\left \{ \left \{ h_{1,d}^{(l)} \right \},\left \{  h_{2,d}^{(l)}\right \} ,\dots, \left \{  h_{N,d}^{(l)}\right \}  \right \} \in\mathbb{R} ^{D \times N}$, where $\left \{  h_{v,d}^{(l)}\right \}  =\left \{ h_{v,1}^{(l)},h_{v,2}^{(l)},\dots ,h_{v,D}^{(l)} \right \} $ is the hidden state set, $v\in V$ is the vertex, $N$ is the number of turbines and $D$ is the number of correlation dimensions. The outputs of Frequency Attention and Feature Attention mechanisms in the $l$-th block are denoted as $H_{ST}^{(l)}\in \mathbb{R} ^{D\times N}$ and $H_{SW}^{(l)}\in \mathbb{R} ^{D\times N}$, respectively, We also denote hidden states of $H_{ST}^{(l)}$ and $H_{SW}^{(l)}$ for vertex $v$ at correlation dimension $d$ as $h_{ST,v,d}^{(l)}$ and $h_{SW,v,d}^{(l)}$, respectively. After CD Gate, we obtain the output of HFF Enhancement Block $H_{O}^{(l)}\in \mathbb{R} ^{D\times N}$.
\paragraph{Frequency Attention:}
Due to the uncertainty of wind power, we designed a Frequency Attention mechanism to capture the dynamic spatial-temporal correlations between wind power and IMFs. The hidden state $h_{ST,v,d}^{(l)}$ for vertices $v \in V$ at correlation dimension $d \in \left \{ 1,2,\dots ,D \right \} $ is denoted as:
\begin{equation}
h_{ST,v,d}^{(l)}=\sum_{v_{i}\in V}\alpha _{ST,v_{i},v}\cdot h_{v,d}^{(l)},
\end{equation}
where $\alpha _{ST,v_{i},v}$ is the attention score indicating the relevance of vertex $ v_{i}$ to $v$ and the summation of attention scores is 1. 

Next, we use the no-masked multi-head attention to calculate the attention scores of $K$ attention heads in parallel. Unlike masked multi-head attention, this approach can globally capture the correlations between IMFs and wind power \citep{liu2023itransformer}. Specifically, as shown in the following equations, we use the dot-product method to compute the relationship between vertices $v_i$ and $v$, and then normalize these scores using the softmax method.
\begin{equation}
z_{ST,v_{i},v}^{(k)}=\frac{\left\langle f_{t,1}^{(k)}(h_{v,d}^{(l)}\left|\right|e^{ST}),f_{t,2}^{(k)}(h_{v,d}^{(l)}\left|\right|e^{ST}) \right \rangle }{\sqrt{\zeta }},
\end{equation}
\begin{equation}
\alpha _{ST,v_{i},v}^{(k)} =\frac{\mathrm{exp} (z_{ST,v_{i},v}^{(k)} )}{ {\textstyle \sum_{v_{i}\in V}} \mathrm{exp} (z_{ST,v_{i},v}^{(k)})}, 
\end{equation}
where $\left |  \right |  $ represents the concatenation operation and $\left \langle \bullet, \bullet  \right \rangle $ denotes the inner product operator, $z_{ST,v_{i},v}^{(k)}$ is the spatial correlation between wind power modes in $k^{th}$ attention head, $\sqrt{\zeta } $ is the scaling factor and $\zeta  = D/K$, where $D$ is the number of correlation states and $K$ is the number of attention heads. We then update the hidden state of each turbine at each dimension as:
\begin{equation}
h_{ST,v,d}^{(l)} =\left |\right | _{k=1}^{K} \left \{ \sum_{v_{i}\in V}\alpha_{ST,v_i,v}^{(k)}\cdot f_{t,3}^{(k)}(h_{v,d}^{(l)})\right \},
\end{equation}
where $\left |\right | _{k=1}^{K}$ indicates the concatenation for each $k$ from 1 to $K$, $f_{t,1}^{(k)} (\bullet )$, $f_{t,2}^{(k)} (\bullet )$ and $f_{t,3}^{(k)} (\bullet )$ represent three different nonlinear projections as:
\begin{equation}
   f(x)=\mathrm{Gelu}(\mathbf{B}x +\mathbf{b}).
\end{equation}
Note that $\mathbf{B} \in \mathbb{R} ^{D\times 1}$ and $\mathbf{b} \in \mathbb{R} ^{D \times 1}$ are learnable parameters and Gelu is the activation function which performs best in Transformer models and avoids the vanishing gradient problem \citep{hendrycks2016gaussian}.
Based on the result of Equation (6), the output of Frequency Attention module at the $l^{th}$ encoder layers is $H_{ST}^{(l)}=\left \{ \left \{ h_{ST,1,d}^{(l)} \right \} ,\left \{  h_{ST,2,d}^{(l)}\right \} ,\dots ,\left \{ h_{ST,N,d}^{(l)} \right \}  \right \} \in \mathbb{R} ^{D\times N}$.

\paragraph{Feature Attention:}
Note that the weather features collected by different turbines are influenced by their respective locations. Therefore, we designed the Feature Attention module to capture the spatial-temporal correlations between weather features and wind power generation. Specifically,  For vertices $v \in V$ at correlation state dimension $d \in \left \{ 1,2,\dots ,D \right \} $, we denote the hidden state $h_{SW,v,d}^{(l)}$ as:
\begin{equation}
h_{SW,v,d}^{(l)}=\sum_{v_{i}\in V}\alpha _{SW,v_{i},v}\cdot h_{v,d}^{(l)},
\end{equation}
where $\alpha _{SW,v_{i},v}$ is the attention score indicating the relevance of vertex $v_{i}$ to $v$ and the summation of attention scores is 1. Similar to Equations (4) and (5), we define the no-masked multi-head attention score $\alpha _{SW,v_{i},v}^{(k)}$, which characterizes the relationship between vertex $v_{i}$ and $v$ for the $k^{th}$ attention head, as follows:

\begin{equation}
z_{SW,v_{i},v}^{(k)}=\frac{\left\langle f_{w,1}^{(k)}(h_{v,d}^{(l)}\left|\right|e^{SW}),f_{w,2}^{(k)}(h_{v,d}^{(l)}\left|\right|e^{SW}) \right \rangle }{\sqrt{\zeta }},
\end{equation}
\begin{equation}
\alpha _{SW,v_{i},v}^{(k)} =\frac{\mathrm{exp} (z_{SW,v_{i},v}^{(k)} )}{ {\textstyle \sum_{v_{i}\in V}} \mathrm{exp} (z_{SW,v_{i},v}^{(k)})}, 
\end{equation}
where $ \left | \right | $ represents the concatenation operation and $\left \langle \bullet, \bullet  \right \rangle $ denotes the inner product operator, $z_{SW,v_{i},v}^{(k)}$ is the spatial correlation between weather features in $k^{th}$ attention head, $\sqrt{\zeta } $ is the scaling factor and $\zeta  = D/K$. We then update the hidden state of each vertex at each dimension as:
\begin{equation}
h_{SW,v,d}^{(l)} =\left |\right | _{k=1}^{K} \left \{ \sum_{v\in V}\alpha_{SW,v_i,v}^{(k)}\cdot f_{w,3}^{(k)}(h_{v,d}^{(l)})\right \},
\end{equation}
where $f_{w,1}^{(k)} (\bullet )$, $f_{w,2}^{(k)} (\bullet )$ and $f_{w,3}^{(k)} (\bullet )$ represent three different nonlinear projections, similar to Equation (8). Finally, the output of Feature Attention module at the $l^{th}$ encoder layer is $H_{SW}^{(l)}=\left \{ \left \{ h_{SW,1,d}^{(l)} \right \} ,\left \{  h_{SW,2,d}^{(l)}\right \} ,\dots ,\left \{ h_{SW,N,d}^{(l)} \right \}  \right \}\in \mathbb{R} ^{D\times N}$.
\paragraph{Correlation Determining Gate (CD Gate):}
The IMFs obtained by VMD contain only part of the temporal information in wind power \citep{dragomiretskiy2013variational}. Simple fusion of $H_{ST}^{(l)}$ and $H_{SW}^{(l)}$ will produce suboptimal results \citep{wang2023weagan}. Therefore, we propose a gated fusion technology to adaptively fuse the spatial-temporal correlations of wind power and weather features across different IMFs. The output of CD Gate module in the $l$-th encoder layer is:
\begin{equation}
 \rho  =\sigma (\mathbf{B}_{ \rho  ,1}H_{ST}^{(l)}+ \mathbf{B}_{ \rho ,2}H_{SW}^{(l)}+\mathbf{b} _{\rho }),
\end{equation}
\begin{equation}
H_{O}^{(l)}= \rho \odot H_{ST}^{(l)}+(1- \rho )\odot H_{SW}^{(l)}.
\end{equation}
where $\mathbf{B} _{\rho ,1}\in \mathbb{R} ^{D\times D}$, $\mathbf{B} _{\rho ,2}\in \mathbb{R} ^{D\times D}$ and $\mathbf{b} _{ \rho }\in \mathbb{R} ^{D \times N}$ are learnable parameters, $\odot $ represents the element-wise product and $\sigma (\bullet )$ is the sigmoid activation function.

\begin{table*}[ht]
\centering
\resizebox{\textwidth}{!}{
\begin{tabular}{c|c|cc|cc|cc|cc|cc|cc|cc}
\toprule
\multirow{2}{*}{Datasets}
& Models 
& \multicolumn{2}{c}{\begin{tabular}[c]{@{}c@{}}\textbf{Hiformer}\\ \textbf{(Ours)}\end{tabular}}   
& \multicolumn{2}{c}{\begin{tabular}[c]{@{}c@{}}ITransformer\\(2024)\end{tabular}} 
& \multicolumn{2}{c}{\begin{tabular}[c]{@{}c@{}}WeaGAN\\(2023)\end{tabular}}  
& \multicolumn{2}{c}{\begin{tabular}[c]{@{}c@{}}RLinear\\(2023)\end{tabular}} 
& \multicolumn{2}{c}{\begin{tabular}[c]{@{}c@{}}D2STGNN\\(2023)\end{tabular}}     
& \multicolumn{2}{c}{\begin{tabular}[c]{@{}c@{}}Autoformer\\(2021)\end{tabular}}
& \multicolumn{2}{c}{\begin{tabular}[c]{@{}c@{}}Informer\\(2021)\end{tabular}}\\ 
\cmidrule(lr){3-4} \cmidrule(lr){5-6} \cmidrule(lr){7-8} \cmidrule(lr){9-10} \cmidrule(lr){11-12} \cmidrule(lr){13-14} \cmidrule(lr){15-16}
& Time steps    
& MAE  & MSE  & MAE & MSE & MAE  & MSE  & MAE  & MSE  & MAE & MSE & MAE & MSE & MAE & MSE\\ 
\toprule
\multirow{5}{*}{\vspace{-0.6cm}\rotatebox{90}{SDWPF}}
&1-h& \textbf{0.18} & \textbf{0.07} &\underline{0.20}&\underline{0.10}& 0.35 & 0.27  & 0.28 & 0.21 & 0.79  & 1.11  & 0.57  & 0.65 & 0.22  & 0.13 \\ \cmidrule{2-16}
&6-h& \textbf{0.19} & \textbf{0.09} &0.32 & 0.25 & 0.42 & 0.40  & 0.50 & 0.56 & 0.80  & 1.11  & 0.70 & 0.89 & \underline{0.30} &\underline{0.21} \\ \cmidrule{2-16}
&12-h& \textbf{0.21} & \textbf{0.11}&0.38 & 0.33 & 0.45 & 0.46  & 0.59 & 0.73 & 0.81  & 1.21  & 0.73 & 1.01 & \underline{0.33} &\underline{0.25} \\ \cmidrule{2-16}
&1-d& \textbf{0.24} & \textbf{0.13} &0.44 & 0.42 & 0.51 & 0.61  & 0.66 & 0.88 & 1.19  & 2.07  & 0.75 & 1.03 & \underline{0.37} &\underline{0.30} \\ \cmidrule{2-16}
&2-d& \textbf{0.27} & \textbf{0.19} &0.49 & 0.50 & 0.55 & 0.65  & 0.73 & 1.08 & 1.21  & 2.44  & 0.81 & 1.24 & \underline{0.44}  &\underline{0.40} \\ 
\toprule
\multirow{5}{*}{\vspace{-0.6cm}\rotatebox{90}{GEFcom}}
&1-h  & \textbf{0.12} & \textbf{0.03} & 0.20 & 0.09 & 0.29& 0.21 & 0.25 & 0.13 &\underline{0.14} & \underline{0.04} & 0.36 & 0.22 & 0.22  & 0.10  \\ \cmidrule{2-16}
&6-h  & \textbf{0.12} & \textbf{0.03} & 0.34 & 0.24 & 0.33& 0.26 & 0.46 & 0.40 &\underline{0.15} & \underline{0.04} & 0.57 & 0.53 & 0.37  & 0.25  \\ \cmidrule{2-16}
&12-h & \textbf{0.13} & \textbf{0.03} & 0.39 & 0.29 & 0.38& 0.30 & 0.59 & 0.60 &\underline{0.23} & \underline{0.08} & 0.58 & 0.54 & 0.43  & 0.34  \\ \cmidrule{2-16}
&1-d  & \textbf{0.15} & \textbf{0.04} & 0.45 & 0.37 & 0.44& 0.40 & 0.70 & 0.80 &\underline{0.22} & \underline{0.07} & 0.73 & 0.82 & 0.49  & 0.43  \\ \cmidrule{2-16}
&2-d  & \textbf{0.16} & \textbf{0.05} & 0.52 & 0.46 & 0.48& 0.43 & 0.79 & 0.98 &\underline{0.22} & \underline{0.07} & 0.79 & 0.94 & 0.55  & 0.51  \\ 
\bottomrule
\end{tabular}
}
\caption{Performance comparison of different models for wind power prediction on SDWPF and GEFcom datasets.}
\label{Table1}
\end{table*}

\begin{figure*}[ht]
\centering
\includegraphics[width=1\textwidth]{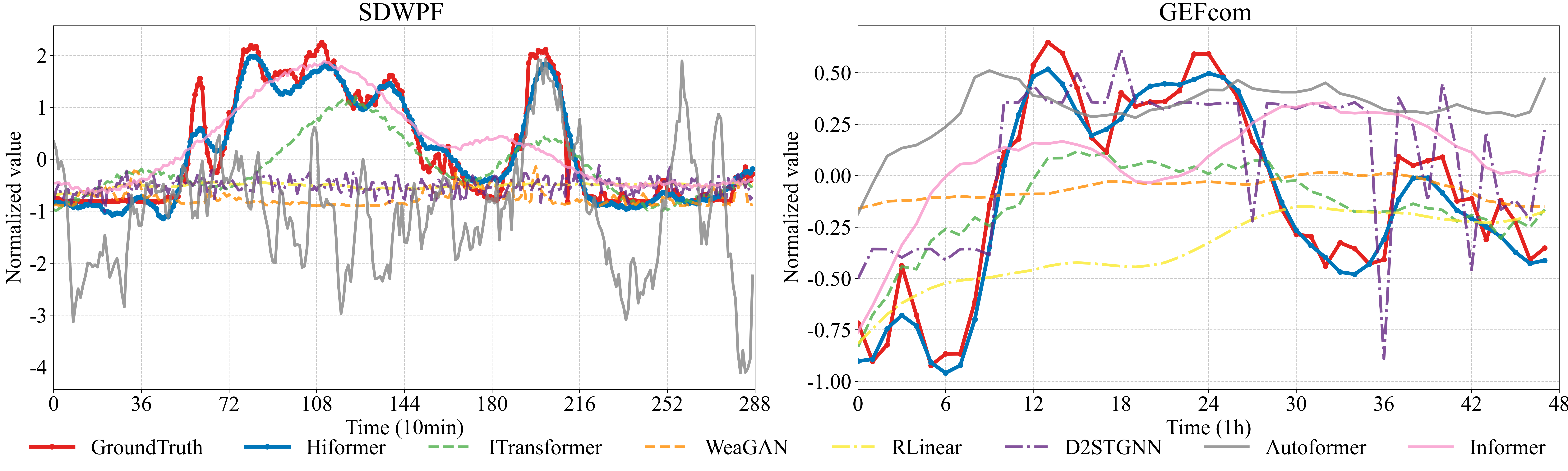} 
\caption{Normalized forecasting results from Aug-7 to Aug-9 across the entire wind farm by all methods for (a) SDWPF dataset and (b) GEFcom dataset.}
\label{Figure4}
\end{figure*}
\subsection{Add$\&$Norm layer}
To increase the convergence and training stability of deep networks \citep{ba2016layer}, we apply normalization to the output of HFF Enhancement block and Feed-Forward layer.
\subsection{Feed-Forward}
We use Feed-Forward network (FFN) to enhance Hiformer's performance of temporal correlation analysis, as commonly done in literature \citep{liu2023itransformer,zeng2023transformers}. The FFN consists of fully-connected layers (FCs) for mapping \citep{wang2023weagan} and a dropout layer to prevent overfitting \citep{srivastava2014dropout}.  The details are shown as below:
\begin{equation}
    H_{F,out}^{(l)}=dropout\left \{ FCs(H_{F,in}^{(l)}) \right \}.
\end{equation}
where $H_{F,out}^{(l)}$ and $H_{F,in}^{(l)}$ is the output and input of FFN at the $l$-th encoder layer, respectively.
\subsection{Projection}
In order to enhance the calculation speed and convergence rate, Hiformer employs Projection module \citep{liu2023itransformer} to decode the encoder's output $H^{(L)} \in \mathbb{R} ^{D\times N}$ and obtain the prediction results $\hat{X}\in \mathbb{R} ^{Q\times N}$. The Projection module primarily consists of MLPs and is represented as follows:
\begin{equation}
    \bm{\hat{X}} =MLPs(H^{(L)}).
\end{equation}

\section{Experiments}
In this section, we conduct empirical studies on Hiformer by applying it to two real-world wind power datasets, demonstrating its prediction accuracy and computational efficiency. Our experiments validated the effectiveness of integrating the signal decomposition technique with weather feature extraction technique.
\subsection{Benchmarks}
We compare Hiformer with the following six Benchmark methods: 
\begin{itemize}
\item $\textbf{ITransformer}$ \citep{liu2023itransformer}: ITransformer adopts inverted attention mechanism and the feed-forward module to achieve long-term prediction.
\item $\textbf{WeaGAN}$ \citep{wang2023weagan}: WeaGAN uses Weather-Aware Graph to capture spatial-temporal correlations.
\item $\textbf{RLinear}$ \citep{li2023revisiting}: RLinear employs Linear Mapping to improve long-term forecasting.
\item $\textbf{D2STGNN}$ \citep{shao2022decoupled}: D2STGNN decouples the time series and proposes a dynamic graph learning module to capture spatial-temporal correlations.
\item $\textbf{Autoformer}$ \citep{wu2021autoformer}: Autoformer decomposes long time series using auto-correlation to enhance the performance in long-term prediction.
\item $\textbf{Informer}$ \citep{zhou2021informer}: Informer uses Generative Style Decoder to get all predictions in one step, which improves the transformer's long-term prediction accuracy and computational efficiency.
\end{itemize}
\subsection{Datasets}
We evaluate the performance of Hiformer on two real-world public wind power datasets: (1) The SDWPF dataset captures various parameters from 134 wind turbines over a span of 430 days, with recordings taken every 10 minutes. The recorded parameters include active power (Patv), reactive power (Prtv), blade pitch angle (Pab), nacelle direction (Ndir), temperature inside the turbine nacelle (Itmp), environment temperature (Etmp), the angle between the wind direction and the nacelle position (Wdir), and wind speed (Wspd) measured by the anemometer \citep{zhou2022sdwpf}; (2) The GEFcom dataset records wind power and wind speed information for 10 zones every hour from 2012 to 2013. The wind speed data is measured at two heights: 10 meters and 100 meters above the ground level \citep{hong2016probabilistic}. 

As it is common in the literature \citep{bubalo2023hybrid,wang2023weagan}, we use the Z-Score method to normalise the  wind power and weather data and split the dataset into training, validation and test sets in a ratio of 7:1:2.
\begin{equation}
    y_{i}=\frac{x_{i}-\bar{x} }{S},
\end{equation}
where $y_{i}$ represents the normalised data, $x_{i}$ is the original data, $\bar{x}$ is the sample mean and $S$ is the sample standard deviation.

\subsection{Experimental Settings}
We have made the code publicly accessible on GitHub anonymously and trained the models using an Intel(R) Xeon(R) Gold 6246R CPU @ 3.40GHz and GeForce RTX 3090 GPUs with PyTorch 1.13. The system is equipped with 128 GB of RAM.

For benchmarks, we use the default settings and best model hyperparameters as specified in their original proposals. We use the 2 days of historical data ($P=288$ time steps for SDWPF and $P=48$ time steps for GEFcom) to predict the wind power for the next $Q\in \left \{ 6,36,72,144,288 \right \}$ time steps in SDWPF and $Q\in \left \{1,6,12,24,48 \right \}$ time steps in GEFcom, following common practices in the literature \citep{ma2023multi,qiu2024novel}. We train our model using the Adam optimizer \citep{kingma2014adam} with an initial learning rate of 0.001. The batch size is set to 64, and the training runs for 100 epochs. The encoder comprises $L=5$ layers, with $K=32$ attention heads and $\zeta =16$ dimensions per attention head, resulting in correlation dimensions $D=K \times \zeta =512$. To prevent overfitting, we set the dropout rate to 0.1. 

We apply two commonly used metrics to evaluate the performance of all the models: Mean Absolute Error (MAE) and Mean Squared Error (MSE) \citep{wang2023weagan}. The equations are as follows:
\begin{equation}
    MAE = \frac{1}{Q}\sum_{t=t_{P+1}}^{t_{P+Q}} \left|(X_{t}-\hat{X} _{t}) \right |,
\end{equation}
\begin{equation}
    MSE = \frac{1}{Q}\sum_{t=t_{P+1}}^{t_{P+Q}} (X_{t}-\hat{X} _{t})^{2}.
\end{equation}
where $\bm{\hat{X}}$ denotes the predicted values and $\bm{X}$ denotes the ground truths.

\subsection{Experimental Results}
\paragraph{Prediction Performance Comparison:}
Table 1 shows the prediction results with the best in bold and the second underlined. As we can see, Hiformer achieves the best prediction results across all time periods, especially for the 2-day prediction length (288 steps in SDWPF and 48 steps in GEFcom). Compared to the second-best method, Hiformer shows improvements of 38.6\% and 52.5\% in MAE and MSE, respectively, over the 2-day horizon in the SDWPF dataset, and 27.3\% and 28.6\% in MAE and MSE, respectively, over the same horizon in the GEFcom dataset. 

To better demonstrate the performance of Hiformer, we visualized the prediction results from two datasets spanning August 7 to August 9. As shown in Figure 4(a), most methods capture the general trend of wind power but struggle with its random fluctuations. This difficulty arises from their inability to account for the strong uncertainty in wind power cased by weather conditions. In contrast, Hiformer effectively captures these fluctuations through the HFF Enhancement block. 

Figure 4(b) shows that most methods perform better on the GEFcom dataset than on the SDWPF dataset. This difference is due to GEFcom has a lower sampling frequency (10 minutes in SDWPF versus 1 hour in GEFcom), a smaller number of turbines (134 turbines in SDWPF versus 10 turbines in GEFcom) and fewer weather features (7 features in SDWPF versus 2 features in GEFcom), which results in lower prediction difficulty.

By comparing the results across these two datasets, we observe that Hiformer exhibits excellent prediction performance across various wind power datasets, especially on tasks involving multiple wind turbines and diverse weather features. This success is attributed to Hiformer’s ability to leverage the advantages of VMD and embedding techniques via the HFF Enhancement block. This combination allows Hiformer to fully explore the correlations between wind power and weather features, resulting in accurate long-term wind power predictions.

\begin{table}[ht]
\centering
\begin{tabular}{c|c|cc}
\toprule
\multirow{2}{*}{Dataset} & \multirow{2}{*}{Models} & \multicolumn{2}{c}{Computation Time} \\
                         &                        & Training(s/epoch)    & Inference(s)   \\ \toprule
\multirow{7}{*}{\rotatebox{90}{SDWPF}} & Informer  &  242.5              &  32.7          \\
                         & Autoformer              &  \underline{220.2}   &  42.9          \\
                         & D2STGNN                 &  390.5              &  45.8          \\
                         & RLinear                 &  245.8              &  36.3          \\
                         & WeaGAN                  &  265.7              &  \underline{22.4}          \\ 
                         & ITransformer            &  245.3              &  34.7          \\ 
                         & \textbf{Hiformer}       &  \textbf{123.0}     &  \textbf{7.4}   \\ 
\toprule   
\multirow{7}{*}{\rotatebox{90}{GEFcom}} & Informer &  220.7              &  23.4          \\
                         & Autoformer              &  \underline{200.1}   &  31.0          \\
                         & D2STGNN                 &  288.6              &  38.7          \\
                         & RLinear                 &  211.9              &  28.9          \\
                         & WeaGAN                  &  249.3              &  \underline{12.1}          \\ 
                         & ITransformer            &  201.7              &  22.1          \\ 
                         & \textbf{Hiformer}       &  \textbf{109.8}     &  \textbf{4.0}   \\ 
\bottomrule
\end{tabular}
\caption{The computation time on the SDWPF and GEFcom datasets}
\label{Table2}
\end{table}

\paragraph{Computation Time:}
Table 2 presents the computation time required by Hiformer and all benchmarks across two datasets with the best in bold and the second underlined. The training phase refers to the average time needed to train the model for one epoch, while the inference phase indicates the total time required for validating the data. As shown in Table 2, Hiformer achieves the best performance in computational time. Compared to the second-best method, Hiformer shows improvements of 44.1\% and 67.0\% in Training time and Inference time, respectively, in the SDWPF dataset, and 45.1\% and 67.0\% in Training time and Inference time, respectively, in the GEFcom dataset. Since D2STGNN decouples the time, it takes the most computation time. Compared to D2STGNN, Hiformer can reduce computation time by up to 68.5\% and 82.8\% in Training time and Inference time, respectively. Although signal decomposition technology typically increases the amount of data and computation time \citep{wang2021review}, Hiformer mitigates this complexity by utilizing an encoder-only structure along with an inverse attention mechanism.

\section{Conclusion}
In this paper, we propose the Hybrid Frequency Feature Enhancement Inverted Transformer (Hiformer) for long-term wind power prediction. By carefully incorporating the Variational Mode Decomposition (VMD) technique within an encoder-only architecture, Hiformer enhances the long-term predictive accuracy for multiple wind turbines while maintaining operational efficiency. Notably, it is the first model to integrate signal decomposition technology with weather feature extraction technique, achieving high-accuracy long-term multi-turbine wind power prediction with low computational time. In future work, we plan to consider the impact of extreme weather events, such as heavy snow and storms, to further improve Hiformer’s prediction accuracy under such conditions.

\bibliography{CameraReady/aaai25}

\end{document}